%% file: main.tex
\documentclass[journal,transmag]{IEEEtran}

\usepackage{amsmath,amsfonts}
\usepackage{algorithmic}
\usepackage{algorithm}
\usepackage{array}
\usepackage[caption=false,font=normalsize,labelfont=sf,textfont=sf]{subfig}
\usepackage{textcomp}
\usepackage{stfloats}
\usepackage{url}
\usepackage{verbatim}
\usepackage{graphicx}
\usepackage{cite}

\usepackage{hyperref}
\usepackage{multirow}
\usepackage{threeparttable}

\usepackage{float} 
\usepackage{graphicx}
\usepackage{caption}
\usepackage{subcaption} 
\usepackage{capt-of}
\usepackage{afterpage}
\usepackage{booktabs}

\begin{document}

\title{A Simple Review of EEG Foundation Models: Datasets, Advancements and Future Perspectives}

\author{
\IEEEauthorblockN{Junhong Lai\IEEEauthorrefmark{1,2,3}, Jiyu Wei\IEEEauthorrefmark{1,2,3}
, Lin Yao\IEEEauthorrefmark{1,2,3} and Yueming Wang\IEEEauthorrefmark{1,2,3,4}
}\\

\IEEEauthorblockA{
\IEEEauthorrefmark{1}The College of Computer Science, Zhejiang University, Hangzhou, China.
}
\IEEEauthorblockA{
\IEEEauthorrefmark{2}The MOE Frontiers Science Center for Brain and Brain-Machine Integration, Zhejiang University, Hangzhou, China. 
}
\IEEEauthorblockA{
\IEEEauthorrefmark{3}The Nanhu Brain-Computer Interface Institute,
Hangzhou, China.
}
\IEEEauthorblockA{
\IEEEauthorrefmark{4}The Qiushi Academy for Advanced Studies, Hangzhou, China.
}

}


\maketitle

\input{chapters/abstract}

\section{Introduction}
\input{chapters/introduction}


\section{Overview of EEG Foundation Models}
    \input{chapters/overview_of_EEGLM/overview}

    \subsection{Data collection and preprocessing}

\input{chapters/overview_of_EEGLM/Data_collection_and_preprocessing}

    \subsection{Model architecture}
    \input{chapters/overview_of_EEGLM/MODEL_ARCHITECTURE}
    \subsection{Reconstruction}
    \input{chapters/overview_of_EEGLM/reconstruction}

\section{Survey Results}
\input{chapters/SURVEY_RESULTS/overview}
    \subsection{Pretraining and Downstream Datasets}
    \input{chapters/SURVEY_RESULTS/datasets_for_pretraining}
    \subsection{Datasets preprocessing}
    \input{chapters/SURVEY_RESULTS/Data_preprocessing}

    \subsection{Parameter Size}
    \input{chapters/SURVEY_RESULTS/parameter_size}
    \subsection{Downstream Task Types}

\input{chapters/SURVEY_RESULTS/downstream_task_evaluation}

    \subsection{Self-Supervise Learning}

\input{chapters/SURVEY_RESULTS/self-supervise}

    \subsection{Hardware for pretraining}

\input{chapters/SURVEY_RESULTS/hardware}

\section{Discussion}
\input{chapters/discussion}

\section{Conclusion}
\input{chapters/conclusion}

\bibliography{main}
\bibliographystyle{unsrt}

\end{document}

%% file: chapters/abstract.tex
\begin{abstract}
Electroencephalogram (EEG) signals play a crucial role in understanding brain activity and diagnosing neurological diseases. Because supervised EEG encoders are unable to learn robust EEG patterns and rely too heavily on expensive signal annotation, research has turned to general-purpose self-supervised EEG encoders, known as EEG-based models (EEG-FMs), to achieve robust and scalable EEG feature extraction. However, the readiness of early EEG-FMs for practical applications and the standards for long-term research progress remain unclear. Therefore, a systematic and comprehensive review of first-generation EEG-FMs is necessary to understand their current state-of-the-art and identify key directions for future EEG-FMs. To this end, this study reviews 14 early EEG-FMs and provides a critical comprehensive analysis of their methodologies, empirical findings, and unaddressed research gaps. This review focuses on the latest developments in EEG-based models (EEG-FMs), which have shown great potential for processing and analyzing EEG data. We discuss various EEG-FMs, including their architectures, pretraining strategies, pretraining and downstream datasets, and other details. This review also highlights challenges and future directions in the field, aiming to provide a comprehensive overview for researchers and practitioners interested in EEG analysis and related EEG-FM.
\end{abstract}

\begin{IEEEkeywords}
Electroencephalogram (EEG), EEG foundation model(EEG-FM).
\end{IEEEkeywords}

%% file: chapters/introduction.tex
\IEEEPARstart{E}{EG} is a non-invasive technique that records the electrical activity of the brain. It has been widely used in neuroscience research, clinical diagnosis, and brain-computer interfaces (BCIs) \cite{li2022eeg}. However, EEG data analysis is challenging due to its low signal-to-noise ratio (SNR), high dimensionality, non-stationarity, and individual variability \cite{jeon2021mutual, du2022eeg}. As highly objective physiological signals, EEG has demonstrated remarkable potential in seizure epilepsy classification \cite{boonyakitanont2020review}, acute stress detection \cite{sharma2022evolutionary}, sleep stage classification \cite{aboalayon2016sleep}, motor imagery recognition \cite{amin2019deep}, abnormal identification \cite{roy2019chrononet}, emotion analysis \cite{suhaimi2020eeg}, and auditory attention detection \cite{biesmans2016auditory}. 

Although EEG modeling has attracted extensive research attention, several issues remain unresolved. Currently, research on EEG recording modeling is mainly divided into two research lines, including manual feature methods \cite{das2020rigor} and deep learning-based methods \cite{wang2022seeg}. Manual feature engineering requires a lot of domain knowledge and may only be applicable to specific tasks. In addition, most deep learning-based methods are fully supervised, which heavily relies on labeled data. However, large-scale labeled data is often infeasible or expensive in medical experiments, highlighting the importance of maximizing labeling efficiency. 

In recent years, the emergence of large-scale pre-training models has revolutionized the field of natural language processing(LLM) \cite{devlin2018bert, qin2024large} and computer vision(VLM) \cite{chen2021empirical, he2022masked, hamadi2023large}. Inspired by these successes, researchers have started to explore the application of large models in EEG analysis \cite{kim2024eeg}. These models can learn hierarchical representations from large amounts of unlabeled EEG data, which can then be fine-tuned for specific tasks, such as seizure detection, emotion recognition, and cognitive state assessment. This approach has the potential to improve the performance and generalization ability of EEG analysis models, while reducing the need for labeled data \cite{zhang2024brant}.

Motivated by the potential of EEG-FMs(EEG Foundation Models) and an increasing number of recent papers proposing EEG-FMs for various EEG tasks, there is an urgent need for a comprehensive review of EEG-FMs for EEG applications. The main contributions of this paper include: 

\begin{itemize}
    \item Summarizing the overview of different EEG-FMs.
    \item Summarizing the datasets for pre-training and downstream of EEG-FMs.
    \item Summarizing the details for pre-training and  evaluation of EEG-FMs.
    \item Future perspectives for EEG-FMs.
\end{itemize}

By addressing these essential aspects, this review paper will provide a comprehensive and in-depth analysis of the application of EEG-FMs for EEG tasks. The goal of this review is to provide a comprehensive overview of the current state-of-the-art EEG foundation models. The models we investigated include the following: Brant\cite{zhang2024brant}\footnote{  \href{https://zju-brainnet.github.io/Brant.github.io/}{https://zju-brainnet.github.io/Brant.github.io/}}, BrainWave(Brant-2)\cite{yuan2024brant}\footnote{  \href{https://github.com/yzz673/Brant-2}{https://github.com/yzz673/Brant-2}}, LaBraM\cite{jiang2024large}\footnote{  \href{https://github.com/935963004/LaBraM}{https://github.com/935963004/LaBraM}}, NeuroGPT\cite{cui2024neuro}\footnote{  \href{https://github.com/wenhui0206/NeuroGPT}{https://github.com/wenhui0206/NeuroGPT}}, BIOT\cite{yangbiot}\footnote{  \href{https://github.com/ycq091044/BIOT}{https://github.com/ycq091044/BIOT}}, EEGPT1\cite{wangeegpt}\footnote{  \href{https://github.com/BINE022/EEGPT}{https://github.com/BINE022/EEGPT}}, BrainBERT\cite{wang2023brainbert}\footnote{  \href{https://github.com/czlwang/BrainBERT}{https://github.com/czlwang/BrainBERT}}, FoME\cite{shi2024fome}\footnote{  \href{https://github.com/1061413241/FoME}{https://github.com/1061413241/FoME}}, EEGPT2\cite{yue2024eegpt}, MBrain\cite{cai2023mbrain}, NeuroLM\cite{jiang2024neurolm}, CBraMod\cite{wang2024cbramod}\footnote{  \href{https://github.com/wjq-learning/CBraMod}{https://github.com/wjq-learning/CBraMod}},  EEGFormer\cite{chen2024eegformer} and ALFEE\cite{xiong2025alfee}\footnote{  \href{https://github.com/wjq-learning/CBraMod}{https://github.com/xw1216/ALFEE}}.

We will discuss the key techniques and architectures used in these models, their performance on various tasks, and the challenges and future directions in this field. By summarizing the latest research findings, we hope to promote further research and development in EEG foundation models and facilitate their application in clinical and research settings.

%% file: chapters/overview_of_EEGLM/overview.tex
EEG-FMs have emerged as a powerful tool in the analysis and interpretation of EEG signals, leveraging large-scale data and unsupervised training approaches, similar to other large models like those used in NLP fields. These models aim to capture complex temporal and spatial dependencies in EEG data, which can be utilized for tasks such as classification, prediction, and anomaly detection. EEG-FMs have gained significant attention due to their ability to model intricate patterns in brain activity, with applications in clinical diagnostics, cognitive neuroscience, BCIs, and more. EEG-FMs typically undergo unsupervised pretraining using vast datasets, allowing the model to learn relevant features and representations from EEG signals without the need for extensive labeled data.

In EEG-related tasks, the input consists of multichannel EEG recordings, where each channel corresponds to a node, and the connectivity between channels (based on functional or structural relationships) represents edges in the model. The data is typically time-series, meaning each EEG signal consists of a series of voltage readings sampled over time, often captured from multiple electrodes placed on the scalp. A large model learns to extract patterns from these signals, and the objective is to predict or classify brain states, such as detecting epileptic seizures, identifying cognitive states, or differentiating between disorders like ADHD or Alzheimer’s disease. 

Compared to traditional machine learning models, EEG-FMs offer several advantages. First, like other large models, they are capable of learning representations directly from raw EEG data, which bypasses the need for manual feature extraction. This ability to handle complex, high-dimensional data makes them particularly powerful for EEG analysis, as they can learn both temporal and spatial patterns from the raw signal. EEG-FMs are also designed to handle the large-scale and diverse nature of EEG data, which may come from different subjects, tasks, and experimental conditions, making them highly versatile and robust. Furthermore, similar to other deep learning models, EEG-FMs can learn hierarchical representations, capturing both low-level features (e.g., oscillatory patterns) and high-level brain network dynamics, leading to improved performance on a variety of tasks. Several advanced techniques in EEG-FMs, such as transformers and self-supervised learning, allow the model to leverage vast amounts of unlabelled EEG data for pretraining, significantly improving generalization to new, unseen data.

The architecture of EEG-FMs varies depending on the task, but common approaches mainly include convolutional neural networks (CNNs) and transformers. These models are typically designed to process the temporal aspect of EEG signals (e.g., using CNNs) as well as the spatial relationships between electrodes (e.g., using attention mechanisms). In particular, transformer models, which have been successful in natural language processing, have been adapted to handle EEG data by processing long-term temporal dependencies and capturing complex spatial relationships between EEG channels. The large-scale nature of EEG-FMs enables them to generalize across different subjects and experimental conditions, making them suitable for a wide range of applications, including brain-computer interfaces (BCIs), cognitive state monitoring, and clinical diagnostics.

EEG-FMs are trained on vast amounts of EEG data from multiple sources, ensuring that the model learns the underlying patterns that are consistent across different individuals and experimental conditions. By using unsupervised learning techniques, such as contrastive learning or self-supervised learning, EEG-FMs can efficiently learn from large datasets without requiring manual labeling, which is often a significant challenge in EEG research. The ability to scale up to large datasets also allows EEG-FMs to adapt to diverse applications, from medical diagnosis to real-time brain-state monitoring in brain-computer interfaces.

%% file: chapters/overview_of_EEGLM/Data_collection_and_preprocessing.tex
Large language model pretraining usually requires a lot of data, and so does the EEG foundation model. Therefore, the first step in pretraining a large model is usually to collect as many EEG datasets as possible to ensure that the diversity and representativeness of the pretraining dataset can fully ensure that the model can fully learn the various characteristics of brain activity.

EEG data usually has a low SNR, and the preprocessing of the data usually includes: bandpass filtering, notch filtering, resampling and normalization. Generally speaking, there is no need to use ICA for additional artifact removal. We discuss the preprocessing steps of each EEG-FM in detail in Section.\ref{section: survey: data preprocessing}.

\begin{itemize}
    \item \textbf{Bandpass filtering}. Bandpass filter the EEG signal to remove low-frequency noise (e.g. 0.1 Hz - 75 Hz).
    \item \textbf{Notch filtering}. Apply a notch filter to avoid power line interference (e.g. 50 Hz).
    \item \textbf{Resampling}. Resample all EEG signals (e.g. 200hz).
    \item \textbf{Normalization}. Normalize the EEG signal (e.g., z-score).
\end{itemize}

%% file: chapters/overview_of_EEGLM/MODEL_ARCHITECTURE.tex
\begin{figure*}[!t]
    \centering
    \includegraphics[width=\linewidth]{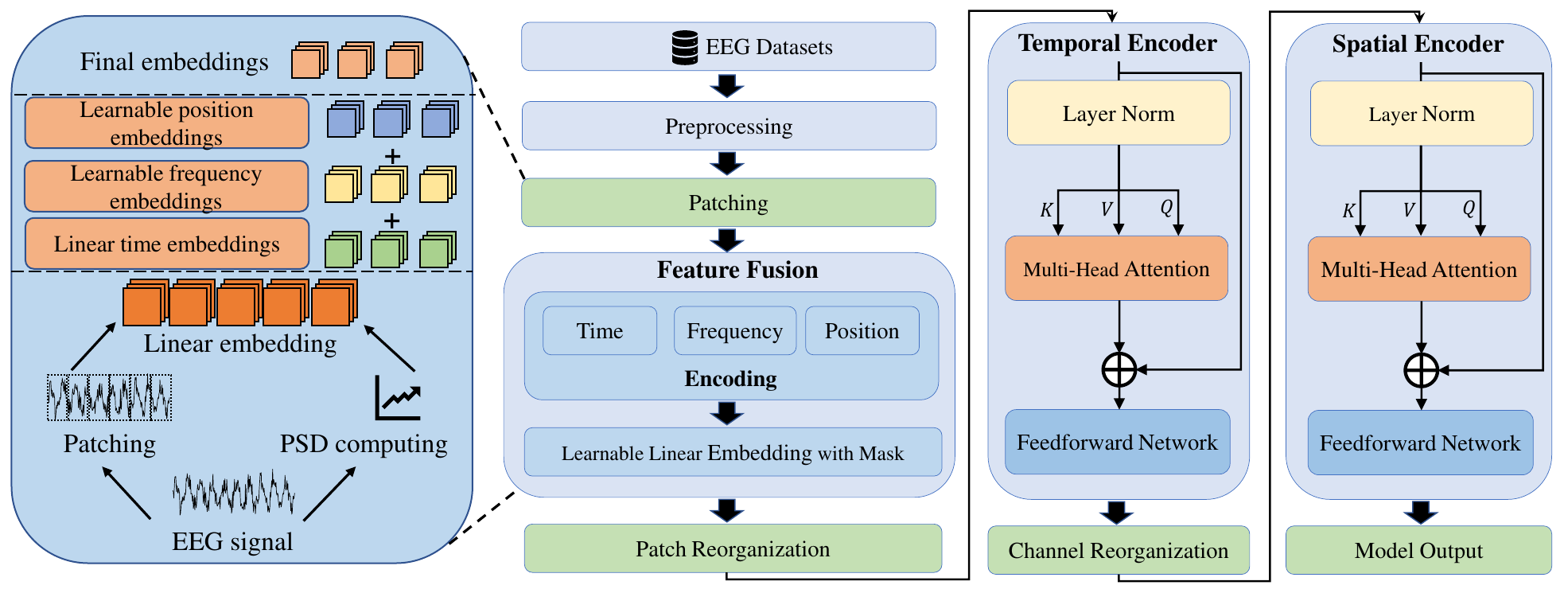} 
    \caption{Overview of the architecture and workflow of general EEG-FM.}
    \label{fig:wide}
\end{figure*}

In this section, we detail the whole framework of universal EEG-FM, though their model architecture details and pre-training details may not be exactly the same. The Overview of the architecture and workflow of general EEG-FM is illustrated in Fig.\ref{fig:wide}. We can formulate the multi-channel EEG signals as $\boldsymbol{X \in \mathbb{R} ^{T \times C}}$, where $C$ is the number of EEG channels and $T$ is the total timestamps.

\subsubsection{Patching}
Since neural recordings are electrical signals with high sampling rates, we aggregate timestamps into patches to (1) enhance the locality and extract semantic information; (2) reduce computation and memory usage; and (3) attend a longer temporal dependency\cite{nie2022time}. Assume that the timestamp for each sample is $t$ and the stride is $s$. $X$ can be segmented into $\left \lfloor \frac{T-t}{s} + 1 \right \rfloor$, and each sample $\boldsymbol{x \in \mathbb{R} ^{t \times C}}$, where $t$ is the number of timestamps and $C$ is the number of electrode channels, we divide $\boldsymbol{x}$ with length $M$, stride $S$ and overlap $O$, where $O = M - S$ is the number of overlapping timestamps between consecutive patches. This generate a set of patches $\boldsymbol{p \in \mathbb{R} ^{N_{p} \times C \times M}}$, where $N_{p} = \left \lfloor \frac{t-M}{S} + 1 \right \rfloor $ is the number of patches in each channel. In fact, $\boldsymbol{p}$ can be represent as:

\begin{equation}
    \left \{  \boldsymbol{p_{k,c}} \in \mathbb{R} ^{M} |k=1,2,...,\left \lfloor \frac{t-M}{S} + 1 \right \rfloor;c=1,2,...,C\right \} 
\end{equation}

\subsubsection{Time-Encoding}
The time-encoding is composed of a linear projection layer. Specifically, the input $\boldsymbol{p \in \mathbb{R} ^{N_{p} \times C \times M}}$ contains $N_{p} \times C$ patches with length $M$ and $\boldsymbol{p_c \in \mathbb{R} ^{N_{p} \times M}}$ denotes the patches in $c$-th channel. We map each sequence of patches $\boldsymbol{p_c \in \mathbb{R} ^{N_{p} \times M}}$ to the latent space of dimension $D$ by a linear projection $\boldsymbol{W_{proj}} \in \mathbb{R} ^{M \times D}$, which can be presented as $e_c = (W_{proj}^{T}p_c)^T$. We denote the time-encoding embeddings as $e \in \mathbb{R} ^{N_{p} \times C \times D}$:

\subsubsection{Frequency-Encoding}
The frequency-encoding mainly extracts the power density value corresponding to each frequency and calculated the power sum within the frequency band according to the preset $n$ frequency band intervals $\left [ f_{min}, f_{max} \right ] $, where $0 < f_{min} < f_{max} < f_{Nyquist} $\cite{landau1967sampling}. For EEG signals, we usually divide the frequency spectrum into five bands: $\delta$ (1-4Hz), $\theta$  (48Hz), $\alpha$  (8-13Hz), $\beta$  (13-30Hz), $\gamma$  (30-50Hz). For the $i$-th frequency band, a learnable encoding $\boldsymbol{f_i} \in \mathbb{R} ^{N_{p} \times D}$ is set as its representation which is shared across all the patches. For a given patch $P_{k,c}(t) = p_{k,c}$, we first apply the Fourier transform \cite{bracewell1989fourier} to convert the time-domain signal into the frequency domain. The power spectral density(PSD), denoted as $P_{k,c}(f)$, is then calculated by squaring the magnitude of the frequency components:

\begin{equation}
    P_{k,c}(f)=\frac{1}{T}\left|\mathcal{F}\{P_{k,c}(t)\}\right|^2=\frac{1}{T}\left|\int_{-\infty}^{\infty}P_{k,c}(t)e^{-i2\pi ft}dt\right|^2
\end{equation}

where $f$ is the frequency, $T$ is the total duration of the signal segment, and $\mathcal{F}$ is the Fourier transform. Subsequently, we compute the sum of power within each sub-band $\left [ f_{min}, f_{max} \right ] $:

\begin{equation}
    PSD_{k,c}(i) = \log_{10}(\sum_{f_i=f_{min}}^{f_{max}}P_{k,c}(f)+1)
\end{equation}

$PSD_{k,c}(i)$ acts as the weight of $\boldsymbol{f_i}$. The frequency encoding $\boldsymbol{F_{k,c}} \in \mathbb{R} ^{D}$ of patch $p_{k,c}$ is obtained as the weighted sum of the learnable encodings $\boldsymbol{f_i}$:

\begin{equation}
    \boldsymbol{F_{k,c}}=\sum_{i=1}^{|\mathrm{bands}|}\frac{\exp(P_{k,c}(i))}{\sum_{i^{\prime}=1}^{|\mathrm{bands}|}\exp(P_{k,c}(i^{\prime}))}\boldsymbol{f_{i}}
\end{equation}

\subsubsection{Position-Encoding}
A learnable positional encoding $\boldsymbol{W_{c, pos}} \in \mathbb{R} ^{N_p \times D}$ monitors the temporal order of patches $\boldsymbol{p_c} \in \mathbb{R} ^{N_{p} \times M}$.

\subsubsection{Time-Frequency Fusion}
According to the previous steps, we can obtain the time-encoding $\boldsymbol{e} \in \mathbb{R} ^{N_{p} \times C \times D}$, frequency-encoding $\boldsymbol{F} \in \mathbb{R} ^{N_{p} \times C \times D}$, position-encoding $\boldsymbol{W_{pos}} \in \mathbb{R} ^{N_p \times C \times D}$. We can obtain the fusion-encoding  $\boldsymbol{\tilde{p}} \in \mathbb{R} ^{N_{p} \times C \times D}$ by adding them together:

\begin{equation}
    \boldsymbol{\tilde{p}} = \boldsymbol{e} + \boldsymbol{F} +\boldsymbol{W_{pos}} 
\end{equation}

\subsubsection{Temporal Encoder}
The Temporal Encoder is designed to capture long-range dependencies and intricate temporal patterns within EEG data, which consists of a stack of dense Transformer encoding layers. Each layer encompasses three key modules: multi-head self-attention, feedforward neural network, and layer normalization with residual connections. Given an input embedding $\boldsymbol{\tilde{p}} \in \mathbb{R} ^{N_{p} \times C \times D}$ obtained from the previous step, we apply a trainable linear transformation along the temporal dimension to derive the query, key, and value matrices for the attention operation, denoted as $\boldsymbol{Q_{time}^i,K_{time}^i,V_{time}^i} \in \mathbb{R} ^{N_{p} \times D}$ respectively. The temporal dependencies are then modeled by computing the attention score $\mathrm{Attention}_{time}^i$:

\begin{equation}
    \mathrm{Attention}_{time}^i = \mathrm{Softmax}(Q_{time}^i(K_{time}^i)^{T}/\sqrt{D} ) V_{time}^i
\end{equation}

In the multi-head self-attention mechanism, the query, key, and value matrices are linearly projected into $D_{k},D_{k},D_{v}$. Subsequently, the attention function is applied to each projected version, yielding an output of dimension $D_{v}$. These outputs are concatenated and mapped back to the original dimension $D$ to obtain the final attention value:

\begin{equation}
    \mathrm{MultiHead} (Q,K,V) = \mathrm{Concat}(head_1, ..., head_h)W^O
\end{equation}

where $head_i = \mathrm{Attention}(QW^Q_i ,KW^K_i, VW^V_i), W^Q_i \in \mathbb{R} ^{D \times D_k}, W^K_i \in \mathbb{R} ^{D \times D_k}, W^V_i \in \mathbb{R} ^{D \times D_v}$, and $W^O \in \mathbb{R} ^{D \times hD_v}$ denote the projection matrix. Finally get the output $\boldsymbol{\hat{p} } \in \mathbb{R} ^{N_{p} \times C \times D}$.

\subsubsection{Spatial Encoder}
The Spatial Encoder is specifically designed to capture spatial relationships and inter-channel dynamics within EEG data. Recognizing the intricate interactions between different brain regions represented by distinct EEG channels, this encoder adapts to learn channel-specific characteristics and their interdependencies. Similarly, given an input embedding $\boldsymbol{\hat{p}} \in \mathbb{R} ^{N_{p} \times C \times D}$ obtained from the previous step, we apply a trainable linear transformation along the channel dimension to derive the query, key, and value matrices for the attention operation, denoted as $\boldsymbol{Q_{ch}^i,K_{ch}^i,V_{ch}^i} \in \mathbb{R} ^{N_{p} \times D}$ respectively. Finally, we compute the attention score $\mathrm{Attention}_{ch}^i$, which involves the interactions between channels and represents the spatial dynamics of the brain:

\begin{equation}
    \mathrm{Attention}_{ch}^i = \mathrm{Softmax}(Q_{ch}^i(K_{ch}^i)^{T}/\sqrt{D} ) V_{ch}^i
\end{equation}

The application of multi-head self-attention mechanism in this part is similar to the previous part. Finally get the output $\boldsymbol{\bar{p} } \in \mathbb{R} ^{N_{p} \times C \times D}$.

%% file: chapters/overview_of_EEGLM/reconstruction.tex
\begin{figure*}[!t]
    \centering
    \includegraphics[width=\linewidth]{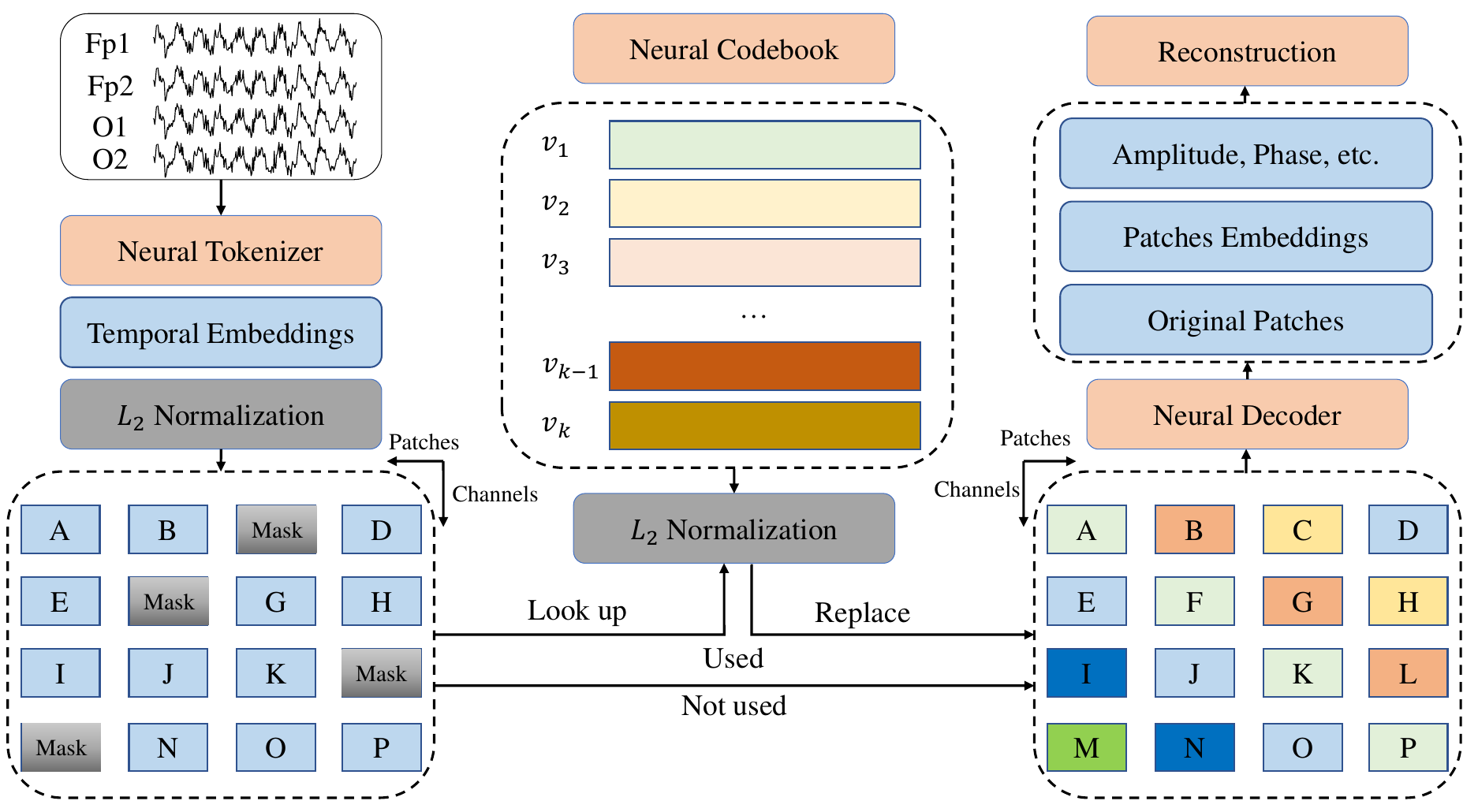} 
    \caption{Overview of neural tokenizer training and EEG-FM pre-training.}
    \label{fig:wide2}
\end{figure*}

Constructing the reconstruction loss allows the model to learn the potential representation of EEG data and the spatio-temporal and frequency domain characteristics of EEG data on unlabeled data through self-supervised learning tasks, thereby improving the model's adaptability to noise and occlusion. The overview of neural tokenizer training and EEG-FM pre-training is illustrated in Fig.\ref{fig:wide2}.

The existing model mainly includes the following four types of reconstruction objectives(see Tab.\ref{tab: reconstruction objectives}): original patches, potential representation of patches(embeddings), frequency domain features of patches, and codebook prediction. 

\subsubsection{Reconstruction of original patches}
During the pre-training stage, the representations $\boldsymbol{\bar{p} } \in \mathbb{R} ^{N_{p} \times C \times D}$ will be fed into a flatten layer with linear head $W_{rec} \in \mathbb{R} ^{D \times M}$ to reconstruct the original patches $\boldsymbol{\breve{p}} \in \mathbb{R} ^{N_{p} \times C \times M}$. Finally we utilize an MSE loss to measure the discrepancy between the reconstructed patches and the original patches:

\begin{equation}
{\mathcal{L}}_{op} = \frac{1}{N_{p}} {\textstyle \sum_{i}^{N_{p}}} ||p_{i} - \breve{p}_{i}||_{2}^{2}
\end{equation}

where $\boldsymbol{\breve{p}_{i},p_{i}  \in \mathbb{R} ^{\times C \times M}}$, $i=1,2,\dots, N_{p}$.

\subsubsection{Reconstruction of patches embeddings}
We utilize an MSE loss to measure the discrepancy between the reconstructed patches embeddings $\boldsymbol{\tilde{e}} \in \mathbb{R} ^{N_{p} \times C \times D}$ and the patches embeddings $\boldsymbol{\tilde{p}} \in \mathbb{R} ^{N_{p} \times C \times D}$:

\begin{equation}
{\mathcal{L}}_{pe} = \frac{1}{N_{p}} {\textstyle \sum_{i}^{N_{p}}} ||\tilde{e}_{i} - \tilde{p}_{i}||_{2}^{2}
\end{equation}

where $\boldsymbol{\tilde{p}_{i},\tilde{e}_{i}  \in \mathbb{R} ^{\times C \times D}}$, $i=1,2,\dots, N_{p}$.

\subsubsection{Reconstruction of fourier spectrum}
For an EEG patch $p_{k,c}=[p[1],p[2],...,p[N_{p}]]$ of channel $c$ and time $k$ in a sample $p$, we apply the Discrete Fourier Transform (DFT) as follows:

\begin{equation}
    \label{eq:DFT}
    {\widetilde{p}}_{c,k}^{m} = \mathop{\sum }\limits_{{n = 1}}^{N_{p}}p\left\lbrack  n\right\rbrack  \exp \left( {-\frac{2\pi j}{N_{p}}{mn}}\right) 
\end{equation}

where $m \in [1,N_{p}]$ and j is the imaginary unit. We rewrite Equation \ref{eq:DFT} using Euler’s formula as:

\begin{equation}
     {\widetilde{p}}_{c,k}^{m} = \mathop{\sum }\limits_{{n = 1}}^{N_{p}}p\left\lbrack  n\right\rbrack  \cos \left( {\frac{2\pi }{N_{p}}{mn}}\right)  - {jp}\left\lbrack  n\right\rbrack  \sin \left( {\frac{2\pi }{N_{p}}{mn}}\right) 
\end{equation}

where ${\widetilde{p}}_{c,k}^{m}$ indicates the spectrum of the sequence at frequency $\omega_{m} = \frac{2\pi m}{N_{p}} $.The the amplitude and phase can be calculated as:

\begin{equation}
\left\{\begin{matrix}
  & {A}^{m} = \sqrt{\operatorname{Re}{\left( {\widetilde{p}}_{c,k}^{m}\right) }^{2} + \operatorname{Im}{\left( {\widetilde{p}}_{c,k}^{m}\right) }^{2}} \\
  & {\phi }^{m} = \arctan \left( \frac{\operatorname{Im}\left( {\widetilde{p}}_{c,k}^{m}\right) }{\operatorname{Re}\left( {\widetilde{p}}_{c,k}^{m}\right) }\right)
\end{matrix}\right. 
\end{equation}

where $Re$ and $Im$ represent the real and imaginary parts of a complex number. We usually apply z-score normalization to normalize ${A}^{m}$ and ${\phi }^{m}$ within a sample for stable convergence.

The output representations are aggregated by average pooling followed by two specific prediction heads to regress the spectrum amplitude $o^A$ and phase $o^\phi$, respectively. The mean squared error (MSE) loss is utilized to guide the prediction:

\begin{equation}
    {\mathcal{L}}_{A} = \frac{1}{N_{p}} \mathop{\sum }\limits_{{i = 1}}^{N_{p}} {\begin{Vmatrix}{o}_{i}^{\phi } - {\phi }_{i}\end{Vmatrix}}_{2}^{2} 
\end{equation}
\begin{equation}
    {\mathcal{L}}_{\phi} = \frac{1}{N_{p}} \mathop{\sum }\limits_{{i = 1}}^{N_{p}}{\begin{Vmatrix}{o}_{i}^{\phi } - {\phi }_{i}\end{Vmatrix}}_{2}^{2} 
\end{equation}

\subsubsection{Codebook prediction}
Codebook prediction consists of two parts:Neural tokenizer training and embedding prediction.

For \textbf{Neural Tokenizer Training}: We define a neural codebook $\mathcal{V} = \left\{  {{v}_{i} \mid  i = 1,\ldots ,K}\right\}   \in  {\mathbb{R}}^{K \times  D}$, where $K$ is the number of the discrete neural embeddings and $D$ is the dimensionality of each embedding. All time-encoding embeddings $e \in \mathbb{R} ^{N_{p} \times C \times D}$ are stored in the codebook $\mathcal{V}$. We utilize a quantizer to quantize all the patch representations into the neural codebook embeddings. The codebook looks up the nearest neighbor of each patch $p_i$ in the neural codebook $\mathcal{V}$. This procedure can be formulated as:

\begin{equation}
    z_i=\arg\min_j\|\ell_2(p_i)-\ell_2(v_i)\|_2,
\end{equation}

where $\ell_2$ represents $\ell_2$ normalization and $z_i$ is the quantized vector after the quantizer. This is equivalent to finding the closest neural embedding by cosine similarity and such $\ell_2$ normalization improves the codebook utilization\cite{peng2022beit}.

\begin{equation}
    {\mathcal{L}}_{ntt} =  \frac{1}{N_p} {\textstyle \sum_{i}^{N_p}}  {\begin{Vmatrix}\mathbf{{sg}}\left( {\ell }_{2}\left( {p}_{i}\right) \right)  - {\ell }_{2}\left( {v}_{{z}_{i}}\right) \end{Vmatrix}}_{2}^{2} + {\begin{Vmatrix}{\ell }_{2}\left( {p}_{i}\right)  - \mathbf{{sg}}\left( {\ell }_{2}\left( {v}_{{z}_{i}}\right) \right) \end{Vmatrix}}_{2}^{2}
\end{equation}

where $sg$ represents the stop-gradient operation that is defined as an identity at the forward pass and has zero gradients.

For \textbf{Embedding prediction}: We randomly generate a mask $\mathcal{M} = \left\{  {{m}_{i} \in  {\left \{ 0,1 \right \} } \mid  i = 1,\ldots ,N_{p}}\right\}  $. After that, we replace the masked patches of $p$ with the learnable mask token $e_\mathcal{M}   \in  {\mathbb{R} ^D}$. The corrupted EEG patches can be denoted as $e^\mathcal{M} = \left \{e_{i}:m_i=0|i=1,2,\dots,N_{p}  \right \} \cup \left \{ e_\mathcal{M}:m_i=1|i=1,2,\dots,N_{p} \right \}$. , which will be added by subsequent modules. $\boldsymbol{\bar{p} } \in \mathbb{R} ^{N_{p} \times C \times D}$ are used to predict the corresponding neural tokens through a linear classifier:

\begin{equation}
    p(\boldsymbol{v}^{\prime}|\boldsymbol{e}^{\mathcal{M}})=\mathrm{softmax}(\mathrm{Linear}(\boldsymbol{\bar{p}}))
\end{equation}

The linear layer outputs a vector $o_i \in \mathbb{R}^\mathcal{V}$ of length $\mathcal{V}$, where each element $o_{i,j}$ represents the unnormalized probability that the patch is predicted to be the $j$-th token in the Codebook. The Softmax function converts the unnormalized probability into a probability distribution.

\begin{equation}
    p(v_j^{\prime}\mid e^{\mathcal{M}})=\frac{\exp(o_{i,j})}{\sum_{k=1}^V\exp(o_{i,k})},\quad j=1,2,\ldots,V
\end{equation}

where $p(\boldsymbol{v}^{\prime}|\boldsymbol{e}^{\mathcal{M}}) $ is the probability that the model predicts that the token corresponding to $\bar{p}_i$ is $v_j$.

The objective training loss is:

\begin{equation}
    \mathcal{L}_{eq}=-\sum_{m_{i}=1}\log p(v_{i}|\boldsymbol{e}^{\mathcal{M}})
\end{equation}


%% file: chapters/SURVEY_RESULTS/overview.tex
This survey is based on a review of 11 articles. These articles were selected by title and abstract screening from a search on Google Scholar and ScienceDirect queried on December 9st, 2024. The search query for collecting the articles was defined as: (“LLM” OR “Foundation model”) AND (“Electroencephalography” OR “EEG”). Both peer-reviewed articles and preprints were included. All types of EEG foundation model were included. 

In the remaining portion of this paper, we report the categories of comparisons we identified in the surveyed papers. These are based on the different aspects of the proposed EEG-based LMs. Specifically, these are:

\begin{itemize}
    \item Datasets for pretraining and downstream tasks.
    \item Preprocessing conditions of datasets.
    \item The evaluation methods of EEG-FMs.
    \item The parameter size of EEG-FMs.
    \item The objective loss function of EEG-FMs.
    \item The hardware conditions for pre-training of EEG-FMs.
\end{itemize}

The following parts will provide further details on these aspects, and the paper will conclude by discussing trends and proposing plausible directions for future research.

%% file: chapters/SURVEY_RESULTS/datasets_for_pretraining.tex
In the pre-training of EEG foundation models, the diversity and representativeness of dataset are crucial. In order to ensure that the model can fully learn the various characteristics of brain activity, a large amount of EEG signal data must be collected. For research teams with conditions, these data not only include public data sets, but can also be combined with self-collected data. Comprehensive data collection usually needs to take into account task diversity, subject diversity, diversity of collection conditions, and diversity of time spans and frequencies. Through comprehensive and diverse data collection, EEG foundation models can obtain richer training data, and then learn more universal and detailed features, thereby demonstrating better performance and generalization capabilities in practical applications. According to our statistics(see Tab.\ref{tab: size and length of datasets}), Brant-2 uses the most pre-training data, reaching 13.79TB, and the EEG recording time is about 40,000 hours.

\begin{table}[h]
\centering
\caption{The size and length of EEG datasets used to pre-train the model}
\begin{tabular}{lcc}
\hline
Model & Size & Length \\
\hline
Brant\cite{zhang2024brant} & 1.01TB & 2528h \\
Brant-2\cite{yuan2024brant} & 13.79TB & 40,907h \\
LaBraM\cite{jiang2024large} & - & about 2,500h \\
NeuroGPT\cite{cui2024neuro} & - & 5656h \\
BIOT\cite{yangbiot} & - & - \\
EEGPT1\cite{wangeegpt} & - & - \\
BrainBERT\cite{wang2023brainbert} & - & 43.7h \\
FoME\cite{shi2024fome} & 1.7TB & about 26,000h \\
EEGPT2\cite{yue2024eegpt} & - & - \\
MBrain\cite{cai2023mbrain} & at least 550GB & at least 470h \\
NeuroLM\cite{jiang2024neurolm} & - & about 25,000h \\
CBraMod\cite{wang2024cbramod} & - & 27,062h \\
EEGFormer\cite{chen2024eegformer} & 1.7TB & about 26,000h \\
ALFEE\cite{xiong2025alfee} & - & about 25,000h \\
\hline
\label{tab: size and length of datasets}
\end{tabular}
    \begin{tablenotes}
    \footnotesize 
    \item[1] - Indicates that the corresponding article does not explicitly mention the size or length of EEG datasets for pre-training.
    \end{tablenotes}
\end{table}

In addition, to evaluate the performance of EEG foundation models in downstream tasks, it is necessary not only to use standard performance indicators, but also to pay attention to the model's generalization ability, interpretability, robustness, computational efficiency and other dimensions. Therefore, the selection of downstream task data also needs to take into account the relevance and complexity of the task. We have collected all relevant datasets for pre-training and downstream tasks for all the models, see the Tab.\ref{tab:Datasets for pretraining and downstream tasks} Tab.\ref{tab:Datasets for pretraining and downstream tasks(continued1)} and Tab.\ref{tab:Datasets for pretraining and downstream tasks(continued2)}. As can be seen from the table, different EEG-FMs utilize varying pre-training datasets and downstream task datasets, which can pose challenges. The size and quality of pre-training data can impact EEG-FM performance, making it difficult to objectively compare the best performance of different EEG-FMs on downstream tasks.

\begin{table}[h]
\centering
\caption{Datasets for pretraining and downstream tasks}
\begin{tabular}{ccc}
\hline
& Pretraining Datasets & Downstream Datasets \\
\hline
\multirow{3}{*}{Brant\cite{zhang2024brant}}
& **        & MAYO\cite{nejedly2020multicenter}  \\
& -        & FNUSA\cite{nejedly2020multicenter}  \\
& -        & **  \\

\hline
\multirow{14}{*}{Brant-2\cite{yuan2024brant}}
& CCEP\cite{van2023developmental}  & AD-65\cite{miltiadous2023dataset}  \\
& CAP\cite{terzano2002atlas}        & CHB-MIT\cite{guttag2010chb}  \\
& HMS\cite{alvarez2021inter}        & Mayo-Clinic\cite{nejedly2020multicenter}  \\
& Siena\cite{detti2020eeg}        & FNUSA\cite{nejedly2020multicenter}  \\
& SRM\cite{hatlestad2022bids}        & MDD-64\cite{mumtaz2016mdd}  \\
& TUEG\cite{harati2014tuh}        & Depression-122\cite{openneuro2021ds003478}  \\
& Schizophrenia-81        & Schizophrenia-28\cite{repod0107441_2017}  \\
& Sleep-EDF\cite{kemp2000analysis} & ADHD-Adult\cite{sadeghibajestani2023eeg}  \\
& Stroke-50\cite{Liu2022}        & ADHD-Child\cite{ieeedata}  \\
& PD-31\cite{rockhill2021uc}        & SD-71\cite{xiang2024resting}  \\
& IowaDataset        & -  \\
& UNMDataset        & -  \\
& AD-184\cite{vicchietti2023computational}        & -  \\
& **       & -  \\

\hline
\multirow{17}{*}{LaBraM\cite{jiang2024large}}
& BCICIV-1\cite{blankertz2007non} & TUAB\cite{obeid2016temple} \\
& Emobrain\cite{savran2006emotion} & TUEV\cite{obeid2016temple} \\
& Grasp and Lift\cite{luciw2014multi} & SEED-V\cite{liu2021comparing} \\
& Inria BCI\cite{margaux2012objective} & MoBI\cite{he2018mobile} \\
& MMI\cite{schalk2004bci2000} & - \\
& RAW\cite{dataverse2020dataset} & - \\
& RSEEG\cite{trujillo2017effect} & - \\
& Siena\cite{detti2020eeg} & - \\
& TVNT\cite{korczowski2019brain} & - \\
& TUAR\cite{buckwalter2021recent} & - \\
& TUEP\cite{veloso2017big} & - \\
& TUSZ\cite{obeid2016temple} & - \\
& TUSL\cite{von2017electroencephalographic} & - \\
& SEED\cite{zheng2015investigating} & - \\
& SEED-V\cite{liu2021comparing} & - \\
& SEED-GER\cite{zheng2018emotionmeter} & - \\
& Self-collected\cite{jiang2023multimodal,jiang2021discriminating,luo2022multimodal,li2021discrimination,tao2020emotion} & - \\

\hline
\multirow{1}{*}{NeuroGPT\cite{cui2024neuro}}
& TUH\cite{obeid2016temple} & BCIC-2A\cite{tangermann2012review} \\

\hline
\multirow{9}{*}{BIOT\cite{yangbiot}}
& SHHS\cite{zhang2018national,quan1997sleep} & -\\
& PREST(*) & - \\
& Cardiology\cite{alday2020classification} & - \\
& CHB-MIT\cite{guttag2010chb} & - \\
& IIIC-Seizure\cite{ge2021deep} & - \\
& TUAB\cite{obeid2016temple} & - \\
& TUEV\cite{obeid2016temple} & - \\
& PTBXL\cite{wagner2020ptb} & - \\
& HAR\cite{anguita2013public} & - \\

\hline

\label{tab:Datasets for pretraining and downstream tasks}
\end{tabular}
    \begin{tablenotes}
    \footnotesize 
    \item[1] ** Indicates the dataset may be private and the authors have not explicitly cited any literature in the paper.
    \item[2] * Indicates the corresponding dataset is private .
    \item[3] The data not cited any literature in the table are not mentioned in the corresponding article.
    \end{tablenotes}
\end{table}

\begin{table}[h]
\centering
\caption{Datasets for pretraining and downstream tasks(continued1)}
\begin{tabular}{ccc}
\hline
& Pretraining Datasets & Downstream Datasets \\

\hline
\multirow{7}{*}{EEGPT1\cite{wangeegpt}}
& PhysioMI\cite{goldberger2000physiobank} & BCIC-2A\cite{tangermann2012review} \\
& HGD\cite{schirrmeister2017deep}      & BCIC-2B\cite{steyrl2016random} \\
& TSU\cite{wang2016benchmark}      & Sleep-EDFx\cite{kemp2000analysis} \\
& SEED\cite{zheng2015investigating}     & KaggleERN\cite{margaux2012objective} \\
& M3CV\cite{huang2022m3cv}     & PhysioP300\cite{goldberger2000physiobank}  \\
& -        & TUAB\cite{obeid2016temple}  \\
& -        & TUEV\cite{obeid2016temple}  \\
\hline

\multirow{8}{*}{FoME\cite{shi2024fome}}
& TUEG\cite{harati2014tuh} & TUEV\cite{obeid2016temple} \\
& SEED\cite{zheng2015investigating} & SEED\cite{zheng2015investigating} \\
& SEED-IV\cite{zheng2018emotionmeter} & SEED-IV\cite{zheng2018emotionmeter} \\
& CHB-MIT\cite{guttag2010chb} & CHB-MIT\cite{guttag2010chb} \\
& Sleep-EDFx\cite{kemp2000analysis} & Sleep-EDFx\cite{kemp2000analysis} \\
& MI-Dataset\cite{schalk2004bci2000} & MAYO\cite{nejedly2020multicenter} \\
& MAYO\cite{nejedly2020multicenter} & FNUSA\cite{nejedly2020multicenter} \\
& FNUSA\cite{nejedly2020multicenter} & - \\
\hline

\multirow{12}{*}{EEGPT2\cite{yue2024eegpt}}
& FACED\cite{chen2023large} & FACED\cite{chen2023large} \\
& SEED\cite{zheng2015investigating}  & DEAP\cite{koelstra2011deap} \\
& SEED-FRA\cite{liu2022identifying} & SEED-IV\cite{zheng2018emotionmeter} \\
& SEED-GER\cite{zheng2018emotionmeter} & SEED-V\cite{liu2021comparing} \\
& SEED-IV\cite{zheng2018emotionmeter} & MIBCI\cite{cho2017eeg} \\
& SEED-V\cite{liu2021comparing} & BCIC4-1\cite{blankertz2007non} \\
& THINGS-EEG-10Hz\cite{grootswagers2022human} & EEGMat\cite{zyma2019electroencephalograms} \\
& THINGS-EEG-5Hz\cite{gifford2022large} & EDF\cite{kemp2000analysis} \\
& IMG(*) & HMC\cite{alvarez2021haaglanden} \\
& - & IMG(*) \\
& - & SPE\cite{nguyen2017inferring} \\
& - & DREAMER\cite{katsigiannis2017dreamer} \\

\hline
\multirow{2}{*}{MBrain\cite{cai2023mbrain}}
& SEEG(*) & SEEG(*) \\
& TUSZ\cite{obeid2016temple} & TUSZ\cite{obeid2016temple} \\
\hline

\label{tab:Datasets for pretraining and downstream tasks(continued1)}
\end{tabular}
    \begin{tablenotes}
    \footnotesize 
    \item[1] ** Indicates the dataset may be private and the authors have not explicitly cited any literature in the paper.
    \item[2] * Indicates the corresponding dataset is private .
    \item[3] The data not cited any literature in the table are not mentioned in the corresponding article.
    \end{tablenotes}
\end{table}

\begin{table}[h]
\centering
\caption{Datasets for pretraining and downstream tasks(continued2)}
\begin{tabular}{ccc}
\hline
& Pretraining Datasets & Downstream Datasets \\
\hline
\multirow{1}{*}{BrainBERT\cite{wang2023brainbert}}
& - & - \\


\hline
\multirow{16}{*}{NeuroLM\cite{jiang2024neurolm}}
& TUEG\cite{harati2014tuh} & - \\
& SEED\cite{zheng2015investigating}  & TUAB\cite{obeid2016temple} \\
& SEED-FRA\cite{liu2022identifying} & TUEV\cite{obeid2016temple} \\
& SEED-GER\cite{zheng2018emotionmeter} & SEED\cite{zheng2015investigating} \\
& SEED-IV\cite{zheng2018emotionmeter} & HMC\cite{alvarez2021haaglanden} \\
& SEED-V\cite{liu2021comparing} & EEGMat\cite{zyma2019electroencephalograms} \\
& BCIC4-1\cite{blankertz2007non} & TUSL\cite{von2017electroencephalographic} \\
& Emobrain\cite{savran2006emotion} & - \\
& Grasp and Lift\cite{luciw2014multi} & - \\
& MMI\cite{schalk2004bci2000} & - \\
& RAW\cite{dataverse2020dataset} & - \\
& RSEEG\cite{trujillo2017effect} & - \\
& Siena\cite{detti2020eeg} & - \\
& SPIS\cite{torkamani2020prediction} & - \\
& TVNT\cite{korczowski2019brain} & - \\
& ** & - \\
\hline

\multirow{12}{*}{CBraMod\cite{wang2024cbramod}}
& TUEG\cite{harati2014tuh} & FACED\cite{chen2023large} \\
& - & SEED-V\cite{liu2021comparing} \\
& - & PhysioMI\cite{goldberger2000physiobank} \\
& - & SHU-MI\cite{goldberger2000physiobank} \\
& - & ISRUC\cite{khalighi2016isruc} \\
& - & CHB-MIT\cite{guttag2010chb} \\
& - & BCIC2020-3\cite{jeong20222020} \\
& - & MDD-64\cite{mumtaz2016mdd} \\
& - & SEED-VIG\cite{min2017driver} \\
& - & EEGMat\cite{zyma2019electroencephalograms} \\
& - & TUEV\cite{obeid2016temple} \\
& - & TUAB\cite{obeid2016temple} \\

\hline
\multirow{5}{*}{EEGFormer\cite{chen2024eegformer}}
& TUEG\cite{harati2014tuh} & TUAB\cite{obeid2016temple} \\
& - & TUAR\cite{buckwalter2021recent} \\
& - & TUSL\cite{von2017electroencephalographic} \\
& - & TUSZ\cite{obeid2016temple} \\
& - & Neonate\cite{stevenson2019dataset} \\

\hline
\multirow{5}{*}{ALFEE\cite{xiong2025alfee}}
& TUEG\cite{harati2014tuh} & TUAB\cite{obeid2016temple} \\
& SEED-IV\cite{zheng2018emotionmeter} & TUAR\cite{buckwalter2021recent} \\
& SEED-V\cite{liu2021comparing} & TUSL\cite{von2017electroencephalographic} \\
& SEED-GER\cite{zheng2018emotionmeter} & TUSZ\cite{obeid2016temple} \\
& SEED-FRA\cite{liu2022identifying} & Neonate\cite{stevenson2019dataset} \\

& BCIC4-1\cite{blankertz2007non} & TUAB\cite{obeid2016temple} \\
& Emobrain\cite{savran2006emotion} & TUEV\cite{obeid2016temple} \\
& Grasp and Lift\cite{luciw2014multi} & SEED\cite{zheng2015investigating} \\
& Inria BCI\cite{margaux2012objective} & HMC\cite{alvarez2021haaglanden} \\
& MMI\cite{schalk2004bci2000} & TUSL\cite{von2017electroencephalographic} \\
& RSEEG\cite{trujillo2017effect} & Workload\cite{zyma2019electroencephalograms} \\
& RawEEG\cite{trujillo2019mental} & - \\
& Siena\cite{detti2020eeg} & - \\
& SPIS\cite{torkamani2020prediction} & - \\
& TVNT\cite{korczowski2019brain} & - \\

\hline
\label{tab:Datasets for pretraining and downstream tasks(continued2)}
\end{tabular}
    \begin{tablenotes}
    \footnotesize 
    \item[1] ** Indicates the dataset may be private and the authors have not explicitly cited any literature in the paper.
    \item[2] * Indicates the corresponding dataset is private .
    \item[3] The data not cited any literature in the table are not mentioned in the corresponding article.
    \end{tablenotes}
\end{table}

%% file: chapters/SURVEY_RESULTS/Data_preprocessing.tex
\label{section: survey: data preprocessing}
In the data preprocessing of EEG signals, denoising, artifact removal and standardization are generally required. If you consider using EEG signals as pre-training data, you generally need to resample all EEG signals to a certain frequency (such as 256hz) and then filter (for example, using filters to remove low-frequency, high-frequency and power noise, such as 0.1Hz and 100Hz bandpass filters and 50Hz notch filters) and then standardize. This process helps to unify the data format and reduce data inconsistency, so that the model can learn features from it more effectively. We have counted the preprocessing of EEG signals for all models, see the Tab.\ref{tab:preprocessing conditions}.

Most EEG-FMs perform minimal and simple data preprocessing steps, namely filtering and resampling, to standardize EEGs from various sources. Notably, the removal of noise-related outliers, suppression of EEG artifacts, artifactual EEG components, and site-related harmonization were not explicitly pursued. Moreover, data normalization strategies that produce training-ready samples were not sufficiently described in most studies. It remains unclear whether or how various offline data handling strategies impact EEG-FM pretraining and downstream task performance.

\begin{table}[t]
\centering
\caption{The preprocessing conditions of EEG datasets}
\begin{tabular}{lccc}
\hline
Model & Resample & Filtering  & Standardize \\
\hline
Brant\cite{zhang2024brant} & $\checkmark$, 250Hz & $\times$ & $\times$ \\
Brant-2\cite{yuan2024brant} & $\checkmark$ & $\checkmark$ & $\times$ \\
LaBraM\cite{jiang2024large} & $\checkmark$, 200Hz & $\checkmark$ & $\checkmark$ \\
NeuroGPT\cite{cui2024neuro} & $\checkmark$, 250Hz & $\checkmark$ & $\checkmark$ \\
BIOT\cite{yangbiot} & $\checkmark$ & $\times$ & $\checkmark$ \\
EEGPT1\cite{wangeegpt} & $\checkmark$, 256Hz & $\checkmark$ & $\checkmark$  \\
BrainBERT\cite{wang2023brainbert} & $\checkmark$ & $\checkmark$ & $\checkmark$ \\
FoME\cite{shi2024fome} & $\checkmark$, 250Hz & $\checkmark$ & $\checkmark$ \\
EEGPT2\cite{yue2024eegpt} &$\checkmark$, 256Hz & $\checkmark$ & $\checkmark$ \\
MBrain\cite{cai2023mbrain} &$\times$ & $\times$ & $\times$ \\
NeuroLM\cite{jiang2024neurolm} &$\checkmark$, 200Hz & $\checkmark$ & $\checkmark$ \\
CBraMod\cite{wang2024cbramod} &$\checkmark$, 200Hz & $\checkmark$ & $\checkmark$ \\
EEGFormer\cite{chen2024eegformer}&$\times$ & $\times$ & $\times$ \\
ALFEE\cite{xiong2025alfee}& $\checkmark$, 256Hz & $\checkmark$ & $\checkmark$  \\
\hline
\label{tab:preprocessing conditions}
\end{tabular}
    \begin{tablenotes}
    \footnotesize 
    \item[1] 1. - Indicates that the corresponding article does not explicitly mention the size or length of EEG datasets for pre-training.
    \item[2] 2. It is worth noting that Brant-2\cite{yuan2024brant} did not uniformly resample the pre-training data in the resampling, which allowed for brain recordings of varying lengths, sampling rates, and electrode counts, enhancing the flexibility of BrainWave for joint pretraining and deployment on iEEG and EEG data. 
    \item[3] 3. BIOT\cite{yangbiot} conducted Ablation Study on Target Resampling Rate.
    \item[4] 4. In Brant-2\cite{yuan2024brant}, BIOT\cite{yangbiot}, BrainBERT\cite{wang2023brainbert} , the resampling frequency is different for different tasks.
    \item[5] 5. MBrain\cite{cai2023mbrain} and EEGFormer\cite{chen2024eegformer} do not mention the description of data preprocessing.
    \end{tablenotes}
\end{table}

%% file: chapters/SURVEY_RESULTS/parameter_size.tex
In the pre-training of EEG-FM, the parameter size of the EEG-FM depends on multiple factors, including the model architecture, the number of layers, the number of neurons in each layer, and the dimension of the input data. In this section, we collect the parameter sizes of all large models, see the Tab\ref{tab: parameter size}.

\begin{table}[t]
\centering
\caption{The parameter size of EEG-FMs.}
\begin{tabular}{lccc}
\hline
Model & Model type  & Size \\
\hline
\multirow{1}{*}{Brant\cite{zhang2024brant}} 
& Brant & 505.69M \\
\hline

\multirow{1}{*}{Brant-2\cite{zhang2024brant}} 
& Brant-2 & - \\
\hline

\multirow{3}{*}{LaBraM\cite{jiang2024large}} 
& LaBraM-Base & 5.8M \\
& LaBraM-Large & 46M \\
& LaBraM-Huge & 369M \\
\hline

\multirow{1}{*}{NeuroGPT\cite{cui2024neuro} } 
& NeuroGPT & 79.53M \\
\hline

\multirow{1}{*}{BIOT\cite{yangbiot}} 
& BIOT & 3.2M \\
\hline

\multirow{2}{*}{EEGPT1\cite{wangeegpt}} 
& EEGPT-Tiny & 4.7M \\
& EEGPT-Base & 25M \\
\hline

\multirow{1}{*}{BrainBERT\cite{wang2023brainbert}} 
& BrainBERT & 43.18M \\
\hline

\multirow{2}{*}{FoME\cite{shi2024fome}} 
& FoME-Base & 476.3M \\
& FoME-Large & 744.8M \\
\hline

\multirow{4}{*}{EEGPT2\cite{wang2023brainbert}} 
& EEGPT-Base & 1.46M \\
& EEGPT-Large & 11.29M \\
& EEGPT-Huge & 183.8M \\
& EEGPT-Giant & 1.09B \\
\hline

\multirow{1}{*}{MBrain\cite{cai2023mbrain}} 
& MBrain & - \\
\hline

\multirow{3}{*}{NeuroLM\cite{jiang2024neurolm}} 
& NeuroLM-B & 254M \\
& NeuroLM-L & 500M \\
& NeuroLM-XL & 1696M \\
\hline

\multirow{1}{*}{CBraMod\cite{wang2024cbramod} } 
& CBraMod & 4.0M\\
\hline

\multirow{1}{*}{EEGFormer\cite{chen2024eegformer} } 
& EEGFormer & -\\
\hline

\multirow{3}{*}{ALFEE\cite{xiong2025alfee}} 
& ALFEE-S & 16.3M \\
& ALFEE-M & 44.3M \\
& ALFEE-B & 120M \\
& ALFEE-L & 300M \\
& ALFEE-XL & 540M \\
\hline

\label{tab: parameter size}
\end{tabular}
    \begin{tablenotes}
    \footnotesize 
    \item[1] - Indicates that the corresponding article does not explicitly mention the parameter size of model.
    \end{tablenotes}
\end{table}

Aside from ALFEE\cite{xiong2025alfee}, few EEG-FM related papers have examined and analyzed the impact of model and data scaling on task performance. Regarding model scaling, most papers lack comprehensive experimental evidence, making it unclear whether a clear trend in model scaling exists. Regarding data scaling, some papers conclude that increasing preprocessed data may lead to poorer performance on downstream tasks or the evidence for data scaling is weak. ALFEE\cite{xiong2025alfee} demonstrates that both model and data scaling apply equally to EEG-FM, with larger models and more preprocessed data generally resulting in better performance.

%% file: chapters/SURVEY_RESULTS/downstream_task_evaluation.tex
The performance of large EEG models is mainly evaluated through several types of downstream tasks, which can be divided into the following categories:

\begin{itemize}
    \item Classification Tasks: These include emotion recognition, seizure detection, cognitive state classification, sleep stage identification, and motor imagery recognition. 
    \item Short-term Prediction: These tasks involve predicting EEG signal trends or cognitive states over short time horizons. 
    \item Long-term Prediction: Long-term prediction tasks focus on forecasting EEG signal patterns or brain state transitions over extended periods. These tasks are essential for applications such as neurodegenerative disease monitoring or long-term cognitive assessment.
    \item Imputation: EEG signal imputation addresses the challenge of missing or corrupted data segments.
\end{itemize}

At the same time, ablation experiments and model analysis are also essential. This section mainly summarizes the downstream tasks and model evaluation methods selected by these EEG-FMs, see Tab\ref{tab:downstream tasks}.

\begin{table*}[]
\centering
\caption{The types of downstream tasks}
\begin{tabular}{l ccccc}
\hline
Model & Classification & Short-term Prediction  & Long-term Prediction & Imputation & Others\\
\hline
\multirow{1}{*}{Brant\cite{zhang2024brant}} 
& Seizure detection 
& \multirow{1}{*}{$\checkmark$}
& \multirow{1}{*}{$\checkmark$}
& \multirow{1}{*}{$\checkmark$}
& Frequency-phase forecasting\\
\hline

\multirow{7}{*}{Brant-2\cite{yuan2024brant}} 
& Alzheimer’s disease diagnosis
& \multirow{7}{*}{$\times$}
& \multirow{7}{*}{$\times$}
& \multirow{7}{*}{$\times$}
& \multirow{7}{*}{-}\\
& Seizure detection\\
& MDD diagnosis\\
& Depression diagnosis\\
& Schizophrenia diagnosis\\
& ADHD diagnosis\\
& Sleep deprivation detection\\
\hline

\multirow{3}{*}{LaBraM\cite{jiang2024large}} 
& Abnormal detection
& \multirow{3}{*}{$\times$}
& \multirow{3}{*}{$\times$}
& \multirow{3}{*}{$\times$}
& \multirow{3}{*}{Regression}\\
& Event type classification\\
& Emotion recognition\\
\hline

\multirow{1}{*}{NeuroGPT\cite{cui2024neuro}} 
& Motor Imagery classification
& \multirow{1}{*}{$\times$}
& \multirow{1}{*}{$\times$}
& \multirow{1}{*}{$\times$}
& \multirow{1}{*}{-}\\
\hline

\multirow{5}{*}{BIOT\cite{yangbiot}} 
& Seizure detection
& \multirow{5}{*}{$\times$}
& \multirow{5}{*}{$\times$}
& \multirow{5}{*}{$\times$}
& \multirow{5}{*}{-}\\
& Event type classification\\
& Seizure type classification\\
& Arrhythmias phenotype prediction\\
& Huamn action recognition\\
\hline

\multirow{6}{*}{EEGPT1\cite{wangeegpt}} 
& Abnormal detection
& \multirow{6}{*}{$\times$}
& \multirow{6}{*}{$\times$}
& \multirow{6}{*}{$\times$}
& \multirow{6}{*}{-}\\
& Event type classification\\
& Motor Imagery classification\\
& Sleep stage detection\\
& Error related negativity detection\\
& Event-related potentials detection\\
\hline

\multirow{4}{*}{BrainBERT\cite{wang2023brainbert}} 
& Sentence onset
& \multirow{4}{*}{$\times$}
& \multirow{4}{*}{$\times$}
& \multirow{4}{*}{$\times$}
& \multirow{4}{*}{-}\\
& Speech vs. non-speech\\
& Volume\\
& Pitch\\
\hline

\multirow{4}{*}{FoME\cite{shi2024fome}} 
& Seizure detection
& \multirow{4}{*}{$\checkmark$}
& \multirow{4}{*}{$\checkmark$}
& \multirow{4}{*}{$\checkmark$}
& \multirow{4}{*}{-}\\
& Sleep stage detection\\
& Emotion recognition\\
& Event type classification\\
\hline

\multirow{5}{*}{EEGPT2\cite{yue2024eegpt}} 
& Emotion Recognition
& \multirow{5}{*}{$\times$}
& \multirow{5}{*}{$\times$}
& \multirow{5}{*}{$\times$}
& \multirow{5}{*}{-}\\
& Motor Imagery classification\\
& Mental workload detection\\
& Sleeping stage detection\\
& Cross Modality\\
\hline

\multirow{2}{*}{MBrain\cite{cai2023mbrain}} 
& Seizure detection
& \multirow{2}{*}{$\times$}
& \multirow{2}{*}{$\times$}
& \multirow{2}{*}{$\times$}
& \multirow{2}{*}{-}\\
& Emotion recognition\\
\hline

\multirow{6}{*}{NeuroLM\cite{jiang2024neurolm}} 
& Abnormal detection
& \multirow{6}{*}{$\times$}
& \multirow{6}{*}{$\times$}
& \multirow{6}{*}{$\times$}
& \multirow{6}{*}{-}\\
& Event Type Classification\\
& Emotion Recognition\\
& Sleep Stage Classification\\
& Workload Detection\\
& Slowing Event Detection\\
\hline

\multirow{10}{*}{CBraMod\cite{wang2024cbramod}} 
& Abnormal detection
& \multirow{10}{*}{$\times$}
& \multirow{10}{*}{$\times$}
& \multirow{10}{*}{$\times$}
& \multirow{10}{*}{-}\\
& Event Type Classification\\
& Emotion Recognition\\
& Sleep Stage Classification\\
& Mental Stress Detection\\
& Vigilance Estimation\\
& Seizure Detection \\
& Imagined Speech Classification \\
& Motor Imagery Classification\\
& MDD Diagnosis\\
\hline

\multirow{4}{*}{EEGFormer\cite{chen2024eegformer}} 
& Abnormal detection
& \multirow{4}{*}{$\times$}
& \multirow{4}{*}{$\times$}
& \multirow{4}{*}{$\times$}
& \multirow{4}{*}{-}\\
& Event Type Classification\\
& Slowing Event Detection\\
& Seizure Detection\\
\hline

\multirow{6}{*}{ALFEE\cite{xiong2025alfee}} 
& Abnormal detection
& \multirow{6}{*}{$\times$}
& \multirow{6}{*}{$\times$}
& \multirow{6}{*}{$\times$}
& \multirow{6}{*}{-}\\
& Event Type Classification\\
& Emotion Recognition\\
& Sleep Stage Classification\\
& Workload Detection\\
& Slowing Event Detection\\
\hline

\label{tab:downstream tasks}
\end{tabular}
    \begin{tablenotes}
    \footnotesize 
    \item[1] 1. MDD stands for Major Depressive Disorder.
    \item[2] 2. ADHD stands for Attention deficit hyperactivity disorder. 
    \end{tablenotes}
\end{table*}

%% file: chapters/SURVEY_RESULTS/self-supervise.tex
Self-supervised representation learning is a powerful approach to extract high level abstract representation from unlabelled data. Among those methods to learn representation via self-supervised pre-training, masked autoencoder (MAE) has been proved to be a simple but effective way in many fields \cite{nie2022time, he2022masked, devlin2018bert}. After preprocessing all datasets involved in pre-training, we further organize the data into multiple signal blocks with approximately equal data volumes. The model treats each signal block as the minimal input unit. This enables cyclical pre-training on multiple datasets. We begin by segmenting the input signal into patches and randomly replacing a certain percentage(e.g. 40\%) of the patches with learnable mask embeddings [MASK]. Then the generalization of the EEG-FM is enhanced by minimizing the reconstruction loss. It should be noted that the reconstruction objectives of different models are different, see the Tab\ref{tab: reconstruction objectives}.

\begin{table}[]
\centering
\caption{The reconstruction objectives of EEG-FM}
\begin{tabular}{lcc}
\hline
Model & Reconstruction Objective & Loss Function\\
\hline
\multirow{1}{*}{Brant\cite{zhang2024brant}} 
& Original patches 
& ${\mathcal{L}}_{op}$\\

\multirow{1}{*}{Brant-2\cite{yuan2024brant}} 
& Embedded tokens
& ${\mathcal{L}}_{pe}$\\

\multirow{2}{*}{LaBraM\cite{jiang2024large}}
& Fourier Spectrum prediction
& \multirow{2}{*}{${\mathcal{L}}_{A}+{\mathcal{L}}_{\phi}+{\mathcal{L}}_{nnt}+{\mathcal{L}}_{eq}$}\\
& Codebook prediction\\

NeuroGPT\cite{cui2024neuro} 
& Embedded tokens 
& ${\mathcal{L}}_{pe}$\\

BIOT\cite{yangbiot} 
& Embedded tokens 
& ${\mathcal{L}}_{pe}$\\

EEGPT1\cite{wangeegpt} 
& Original patches 
& ${\mathcal{L}}_{op}$\\

BrainBERT\cite{wang2023brainbert} 
& Masked spectrogram 
& ${\mathcal{L}}_{pe}$\\

FoME\cite{shi2024fome} 
& Original patches 
& ${\mathcal{L}}_{op}$\\

EEGPT2\cite{yue2024eegpt}
& Original patches 
& ${\mathcal{L}}_{op}$\\

MBrain\cite{cai2023mbrain} 
& - 
& - \\

\multirow{3}{*}{NeuroLM\cite{jiang2024neurolm} }
& Original patches
& \multirow{3}{*}{${\mathcal{L}}_{op}+{\mathcal{L}}_{\phi}+{\mathcal{L}}_{nnt}+{\mathcal{L}}_{eq}$}\\
& Frequency magnitude\\
& Codebook prediction\\

CBraMod\cite{wang2024cbramod}
& Original patches 
& ${\mathcal{L}}_{op}$\\

\multirow{2}{*}{EEGFormer\cite{chen2024eegformer} }
& Original patches
& \multirow{2}{*}{${\mathcal{L}}_{op}+{\mathcal{L}}_{nnt}+{\mathcal{L}}_{eq}$}\\
& Codebook prediction\\

\multirow{3}{*}{ALFEE\cite{xiong2025alfee} }
& EEG signal forecasting
& \multirow{3}{*}{see \cite{xiong2025alfee}}\\
& Original patches\\
& Task classification
\\

\hline
\label{tab: reconstruction objectives}
\end{tabular}
    \begin{tablenotes}
    \footnotesize 
    \item[1] 1. - Indicates that the corresponding article does not explicitly mention the reconstruction details of EEG-FM.
    \item[2] 2. In LaBraM\cite{jiang2024large}, the loss fails to converge while directly reconstructing raw EEG signals(Original patches).
    \item[3] 3. NeuroLM\cite{jiang2024neurolm} observes that reconstructing the phase contributes minor to neural tokenizer training.
    \end{tablenotes}
\end{table}

%% file: chapters/SURVEY_RESULTS/hardware.tex
In this section we summarize the hardware conditions for the pre-training of EEG-FM, see the Tab.\ref{tab:hardware conditions}.

\begin{table}[]
\centering
\caption{The hardware conditions for the pre-training of EEG-FM}
\begin{tabular}{lccc}
\hline
Model & GPU & Time & Quantity \\
\hline
Brant\cite{zhang2024brant} & NVIDIA Tesla A100 & 2.8 days & 4\\
Brant-2\cite{yuan2024brant} & NVIDIA Tesla A100 & 100 hours & 4\\
LaBraM\cite{jiang2024large} & NVIDIA A800  & - & - \\
NeuroGPT\cite{cui2024neuro} & - & - & - \\
BIOT\cite{yangbiot} & NVIDIA RTX A6000 & - & 8 \\
EEGPT1\cite{wangeegpt} & NVIDIA RTX 3090 & - & 8  \\
BrainBERT\cite{wang2023brainbert} & - & - & - \\
FoME\cite{shi2024fome} & NVIDIA RTX 4090 & 350 hours & 6 \\
EEGPT2\cite{yue2024eegpt} & NVIDIA A800-SXM4 & 20 hours & 8 \\
MBrain\cite{cai2023mbrain} & - & - & - \\
NeuroLM\cite{jiang2024neurolm} & NVIDIA Tesla A100 & - & 8 \\
CBraMod\cite{wang2024cbramod} & NVIDIA RTX A5000 & 5 days &  4\\
EEGFormer\cite{chen2024eegformer} & - & - & - \\
ALFEE\cite{xiong2025alfee} & NVIDIA A800 & - & 8 \\
\hline
\label{tab:hardware conditions}
\end{tabular}
    \begin{tablenotes}
    \footnotesize 
    \item[1] - Indicates that the corresponding article does not explicitly mention the hardware conditions for pre-training.
    \end{tablenotes}
\end{table}

%% file: chapters/discussion.tex




\subsection{Current Research Challenges}
\textbf{Effects of Preprocessing and Normalization}: EEG datasets may require extensive offline preprocessing to manage data quality and suppress artifacts. Because the current study performed minimal processing, it is unclear whether data outliers and noise impact EEG-FM preprocessing and task performance. Even clean EEG datasets require careful consideration of normalization strategies, as EEG can vary across channels, subjects, and acquisition sites\cite{Wagh2022}. Therefore, there is a need to better understand the impact of these choices on downstream EEG-FM modeling.

\textbf{Standardized Benchmark Tasks}: The lack of common tasks in EEG-FM evaluation makes understanding the state of the art challenging and highlights the need to identify a common core evaluation set for future EEG-FM evaluation. This set must cover a variety of task types, including classification and regression, and have both dense (one label for the entire EEG recording) and sparse (one label for each EEG segment) labels. Furthermore, some existing datasets may have saturated the performance of traditional methods\cite{kiessner2024reaching}, preventing them from fully exploring the potential of EEG-FM. Therefore, collecting and open-sourcing challenging new datasets is crucial for the future development of EEG-FM.

\textbf{Long-Term Context Modeling}: The required EEG recording duration depends on the signal-to-noise ratio of the target neural signal, the statistical requirements for improving the signal-to-noise ratio through repeated sampling, and the temporal characteristics of the brain activity being studied.These include:
\begin{itemize}
    \item Transient events (milliseconds to seconds): ERPs\cite{WikipediaERP, zang2022eeg, navajas2013uncovering, tian2018classification}, such as the P300 (peak approximately 300 ms after stimulus), the N170 (peak approximately 170 ms), and the error-related negativity (ERN, peak approximately 100 ms after the response), are brief, rapid responses to discrete stimuli or events. The "unit of analysis" here is a short time window (epoch) surrounding the event, typically 1-2 seconds.
    \item Rhythmic states (seconds to minutes): Resting-state or task-state\cite{zhao2025electrophysiological} EEG analysis focuses on persistent EEG oscillations that remain relatively stable over several minutes, such as alpha, beta and specific task-related rhythms. In these cases, the "unit of analysis" is a continuous recording lasting several minutes.
    \item Evolutionary states (tens of minutes to hours): Cognitive states\cite{mastropietro2023reliability, so2017evaluation, li2025effects, di2018eeg}, such as driving fatigue and mental workload, evolve gradually during task performance. The "unit of analysis" here is the duration of the entire task, as the study is precisely about changes in states over time.
    \item Macrostructural Patterns (Hours): Sleep staging\cite{chou2011minimum, zhou2024simplifying} aims to observe the complete structure of sleep, including multiple approximately 90-minute cycles of non-rapid eye movement (NREM) and rapid eye movement (REM) sleep. This requires a full night of recording, typically 6-8 hours. The "unit of analysis" is the entire sleep period.
\end{itemize}

\textbf{Trustworthy Modeling}: Currently, no EEG-FMs focus on model interpretability or explanation, which remains a key requirement for data-driven modeling in high-stakes and expert-centric medical fields. Research is needed to uncover the black box of EEG-FMs and gain insights into the actual robustness of their decision-making processes, both in pre-training and in downstream applications. Linking to known brain physiological or pathological patterns may be necessary for expert users to better assist in EEG-based diagnosis of disorders (such as ASD, ADHD, and MDD).

\subsection{Performance Evaluation}

One of the most significant obstacles in current EEG-FM research is the lack of standardized evaluation protocols and benchmarks\cite{kuruppu2025eegfoundationmodelscritical}. Different studies evaluate their models on a wide range of downstream tasks and datasets, making direct and fair comparisons nearly impossible. For example, one model may perform well on a seizure detection task, while another performs poorly on an emotion recognition task. Even for the same task (e.g., motor imagery), the datasets, preprocessing pipelines, and data partitioning strategies (e.g., within-subject vs. between-subject cross-validation) used can vary, making performance results incomparable. This heterogeneity hinders a clear understanding of true progress in the field. To address this issue, the research community urgently needs to develop and adopt standardized benchmark suites, such as emerging initiatives like AdaBrain-Bench\cite{wu2025adabrain}\footnote{\href{https://github.com/Jamine-W/AdaBrain-Bench}{https://github.com/Jamine-W/AdaBrain-Bench}}, which aims to provide a framework covering a variety of BCI tasks, multi-dimensional evaluation metrics, and standardized adaptation procedures. Such benchmarks will provide a level playing field for evaluating generalization across subjects, tasks, and datasets.

Many models are evaluated under flawed settings. A common problem is "in-sample" evaluation, where the downstream task data used for fine-tuning and testing is itself part of the pre-training dataset. This approach fails to truly test the model's generalization ability to new, unseen data. A more rigorous and realistic evaluation metric is "out-of-distribution" (OOD) evaluation, where model performance is tested on datasets completely unseen during model pre-training. This simulates real-world scenarios when deploying the model to a new hospital or using it with new hardware. Furthermore, the evaluation should clearly distinguish between cross-subject generalization (i.e., performance on new subjects in the same dataset) and the more challenging cross-dataset generalization (i.e., performance on datasets from different studies with different protocols and hardware). Only through rigorous OOD and cross-dataset testing can we truly trust that the model has learned universal neurophysiological representations, rather than biases or artifacts from a specific dataset.

In addition, methods for evaluating EEG-FM capabilities also need to diversify. Currently, the dominant evaluation paradigm is to fully fine-tune the entire pre-trained model on a downstream task. While this demonstrates the model's best potential performance, it does not fully reveal the intrinsic quality of the representations learned during pre-training\cite{BrainAccess2025}. 
\begin{itemize}
    \item Linear probing: In this strategy, the pre-trained encoder weights are frozen, and only a simple linear classifier is trained on top of them. The performance of linear probing directly reflects the quality and separability of the learned representations. However, many studies have reported that linear probing performs significantly worse than full fine-tuning, raising questions about the intrinsic quality of representations learned by current self-supervised learning strategies. 
    \item Few-Shot and Zero-Shot Learning: This is a critical test of the capabilities of a base model. Evaluating a model's ability to handle new tasks with only minimal (few-shot) or even no (zero-shot) labeled training examples is central to measuring its true utility as a general-purpose feature extractor.
\end{itemize}

Finally, evaluation metrics themselves need to go beyond traditional classification accuracy to better reflect the model's actual clinical or application value. For example, in epilepsy detection tasks, false positives per hour is a more clinically relevant metric than the F1 score because it directly relates to alarm fatigue and clinical workflow\cite{saab2024towards}. Similarly, when assessing continuous states such as stress or fatigue, models must not only be accurate but also capable of early prediction\cite{jaiswal2022assessing}, which requires specialized evaluation metrics. Therefore, future evaluation frameworks should integrate more application-specific and meaningful metrics to comprehensively assess a model's real-world utility.



\subsection{Future Directions}
Although great progress has been made in the research of EEG foundation models, several promising directions can still be found in the rapidly developing field of EEG-FM research. 

First, almost all current EEG-FMs use a self-supervised learning framework with masks. The key operation to achieve this is to segment the EEG signal into blocks, inspired by block embeddings in images. A comprehensive comparison of the various detailed parameters of EEG segmentation (e.g. the architecture settings of EEG-FM, the sampling rate of the EEG signal, the length of the EEG patches, the overlap rate between the EEG patches, the percentage of data masked in self-supervised training and the selection of pre-training datasets) with respect to their influence on downstream tasks performance should be carried out to address this crucial design question in a systematic manner. Only articles \cite{cui2024neuro, yangbiot} reported the impact of various detailed parameters of EEG segmentation on the experimental results: Article\cite{cui2024neuro} mentioned that increasing the length of patches, increasing the number of patches, reducing the overlapping ratio, increasing the embedding dimension, and increasing the number of self-attention layers in the encoder can enhance the model performance of downstream tasks, the article does not provide convincing data and charts to prove its conclusions. Article\cite{yangbiot} mentioned that increasing the length of blocks, the sampling rate of the EEG signal and increasing the overlapping ratio can enhance the model performance of downstream tasks with detailed ablation experiments. However, their views differ in the overlapping ratio.

Secondly, GNN can be considered to capture the spatial pattern of EEG signals. We found that EEGPT2 \cite{yue2024eegpt}  and \cite{wang2024graph} has tried to integrate GNN into EEG-FM to better capture the potential spatial features of EEG signals. Thirdly, the number of parameters of EEG-FM is still small compared to the current mainstream large language models(e.g. ChatGPT-4o). How to design models that fit specific tasks well but have small parameters, and how to design models with large parameters and strong generalization will be the focus of future work. Fourthly, designing a model that generates EEG signals from language will be a competitive direction, which may solve the problem of lack of model pre-training EEG signals. Finally, current work primarily focuses on robust EEG representation, further research is needed to extend the model to multimodal scenarios such as Video-EEG, Image-EEG, and Text-EEG. Therefore, combining EEG models with other modes such as language and imaging to create multimodal vertical fields EEG-FM is expected to create an overall framework connecting neuroscience and artificial intelligence, opening up new avenues for research and application.

%% file: chapters/conclusion.tex
EEG-FMs have become a powerful tool for analyzing and understanding EEG signals. Through their advanced architectures and pre-training strategies, these models have achieved superior performance in various tasks, including disease diagnosis, sleep monitoring, and emotion classification. In this article, we summarize various details of EEG-FMs, including pre-training techniques, pre-training datasets, downstream task datasets, etc. As EEG-based models continue to advance, their transformative impact on healthcare, neuroscience, and brain-computer interface technology is expected to grow, bridging the gap between theoretical progress and practical applications. 

In summary, EEG-FMs are expected to improve our understanding of the brain and develop new applications in neuroscience and healthcare. Continued research and innovation in this field will help develop more accurate, efficient, and interpretable EEG-FM, ultimately achieving better diagnosis, treatment, and rehabilitation of neurological diseases.